# Detecting and Mitigating Algorithmic Bias in Binary Classification using Causal Modeling

WENDY W. Y. HUI*

Singapore Institute of Technology

Email: wendy.hui@singaporetech.edu.sg

WAI KWONG LAU

University of Western Australia

Email: john.lau@uwa.edu.au

This paper proposes the use of causal modeling to detect and mitigate algorithmic bias. We provide a brief description of causal modeling and a general overview of our approach. We then use the Adult dataset, which is available for download from the UC Irvine Machine Learning Repository, to develop (1) a prediction model, which is treated as a black box, and (2) a causal model for bias mitigation. In this paper, we focus on gender bias and the problem of binary classification. We show that gender bias in the prediction model is statistically significant at the 0.05 level. We demonstrate the effectiveness of the causal model in mitigating gender bias by cross-validation. Furthermore, we show that the overall classification accuracy is improved slightly. Our novel approach is intuitive, easy-to-use, and can be implemented using existing statistical software tools such as *lavaan* in R. Hence, it enhances explainability and promotes trust.

## 1 INTRODUCTION

Amid the increasing prevalence of AI in business and industries, ensuring unbiased outcomes by mitigating inherent biases and enhancing AI's explainability is essential to maintain equity and promote trust among various stakeholders. This paper presents a post-processing approach to AI fairness using causal modeling to: (1) detect algorithmic biases and (2) provide statistical remedies to correct the biases.

In the context of AI fairness, we are concerned with biases that are based on protected attributes such as gender, race, religion, etc. [1]. Our goal is to ensure that outcomes from AI do not exhibit any bias based on these attributes. To demonstrate AI fairness, the most commonly used approach is to establish statistical parity. This approach attempts to equalize or maintain an acceptable level of disparity in the output from AI between protected and non-protected groups. For example, Pymetrics, a company that uses AI to evaluate job applicants, mitigates bias in their AI models by ensuring that the selection rate for any protected group is at least 80% of the selection rate of the non-protected

---

group, in accordance with US' Equal Employment Opportunity Commission [2, p. 42]. However, for certain job categories, some of the protected attributes may correlate with job performance, e.g., jobs that require physical strength. Relying solely on statistical parity may lead to "positive discrimination", which may undermine the principle of meritocracy. According to this principle, employers should recruit employees based on merit, regardless of age, race, gender, religion, marital status, family responsibilities or disability [3].

Our proposed approach to AI fairness involves creating a causal model of biases; determining the statistical significance of the biases; and providing statistical remedies to correct for the biases. The advantages of our proposed approach are as follows:

1. The proposed approach allows protected attributes to correlate with the underlying factor that forms the basis of decisions (e.g., job performance). As a result, it can prevent the unintentional introduction of positive discrimination in AI solutions.
2. The statistical nature of the proposed approach means that the results are easily interpretable and can be used to explain the nature of the biases introduced by an AI application and how the biases are corrected, thereby enhancing explainability and promoting trust among different stakeholders of AI.
3. Our approach separates AI development from bias detection and correction, eliminating the need for de-biasing during the training or pre-training phase and significantly saving time.

In the remainder of the paper, we will review some related work in AI fairness, describe our research method, present our experiments, evaluate the effectiveness of our proposed method, summarize the paper and identify a number of future research directions.

## 2 RELATED WORK

There are three approaches to reducing arithmetic bias: (1) pre-processing, (2) in-processing and (3) post-processing [4]. Pre-processing involves manipulating data before training the AI model. For example, some studies have attempted to address the problem of data imbalance by augmenting the training data using synthesized data [4-6]. The in-processing approach, on the other hand, incorporates fairness in the training algorithm. For example, [7] employs adversarial training to maximize prediction accuracy while making it difficult for the adversary to determine the protected attributes. The method proposed in [8] incorporates both pre-processing and in-processing by using generative augmentation to address data imbalance and adversarial training to maximize independence between predictions and protected attributes.

On the other hand, post-processing treats the AI model as a black box and aims to reduce bias by manipulating the output from AI. Focusing on binary classification, [9] demonstrates how fairness can be achieved by equalized odds or equal opportunity, and thus allowing the protected attributes to correlate with the target. [10] extends [9] to provide the solution for establishing multiclass fairness. Instead of equalizing odds or opportunity, our proposed approach makes use of causal models to detect bias and mitigate bias.

The use of causal models in AI fairness is relatively new. Two exceptions are [11] and [12]. Madras et al. [11] analyzed how causal models can improve prediction accuracy in the presence of confounding factors using experiments based on health data. Khademi et al. [12] demonstrated how causal models can be used to detect biases using a synthesized data set and two publicly available data sets. However, in addition to bias detection, we will demonstrate how causal models can be used to correct for biases. To the best of our knowledge, we are the first to apply causal modelling in post-processing for AI fairness.

## 3 METHOD

We denote the protected attributes as $a_1, a_2, a_3$, etc. In real life, these attributes typically include gender, age, religion, ethnicity, and so on. The non-protected attributes are denoted as $x_1, x_2, x_3$, etc. An example of a non-protected attribute is the number of years of academic qualification. The target is represented as $y$, while the predicted value of the target is denoted as $\hat{y}$. $y$ could be the output or revenue generated by an employee. In classification problems, $y$ is categorical. For instance, it could represent the promotion decision with two possible values: "promote" and "not promote." Like [9], this paper focuses on binary classification. Classification models often produce probabilities or real-valued scores as output, and hence, we assume $\hat{y}$ to be continuous.

Figure 1 shows the conceptual model of this study. The prediction model takes the protected and unprotected attributes as input and outputs the prediction $\hat{y}$. Here, the prediction model is treated as a black box. To determine if there is any bias in $\hat{y}$, a causal model is created based on the protected attributes, the target $y$, and the predicted $\hat{y}$, as enclosed by the dotted rectangle.

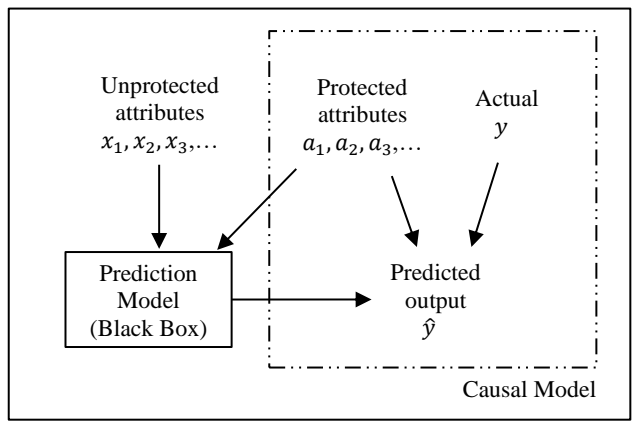

Figure 1: Conceptual Model

Causal modelling is an extension of regression and allows for more than one endogenous (i.e., dependent) variable in the model. Furthermore, the technique can be used to incorporate unobservable, latent variables [13]. A causal model is typically presented using a path diagram, as shown in Figure 2. For simplicity, we consider only one protected attribute $a$. Observable variables are depicted as rectangles, while latent (i.e., unobservable) variables are depicted as circles or ovals. In this case, we have an error term $e$ as a latent variable and we assume that it has a mean of zero. A constant term is represented as a triangle. In this case, we use the constant term one to model the intercept term. The arrows represent causal relationships. In Figure 2, $a$ affects $\hat{y}$, but not the other way round. Similarly, $a$ affects $y$ and $y$ affects $\hat{y}$. Each arrow is associated with a path coefficient, which indicates the strength of the causal relationship.

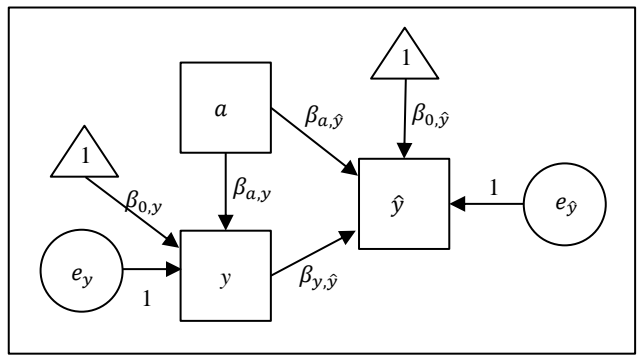

Figure 2: Our Causal Model

Taken together, the mathematical equations that describe the model are:

$$\begin{cases} \hat{y} = \beta_{0,\hat{y}} + \beta_{a,\hat{y}}a + \beta_{y,\hat{y}}y + e_{\hat{y}} & (1) \\ y = \beta_{0,y} + \beta_{a,y}a + e_y & (2) \end{cases}$$

Given sufficient data, existing causal modeling techniques can be applied to estimate the intercept terms ($\beta_{0,\hat{y}}$ and $\beta_{0,y}$) and the path coefficients ($\beta_{a,y}$, $\beta_{y,\hat{y}}$, and $\beta_{a,y}$). In the absence of bias, $\hat{y}$ should only be dependent on $y$. All effects of $a$ should be mediated through $y$. Hence, the path coefficient of the path from $a$ to $\hat{y}$ ($\beta_{a,\hat{y}}$) should be zero. If $\beta_{a,\hat{y}}$ is significantly different from zero, there is sufficient evidence to suggest that the prediction model has introduced bias. In that case, based on the developed causal model, we can estimate what $\hat{y}$ should be in the absence of the bias by setting $\beta_{a,\hat{y}} = 0$. This will be the bias-mitigated value of $\hat{y}$. Let's call it $\tilde{y}$. Decisions based on $\tilde{y}$ will demonstrate improved algorithmic fairness, as we shall show in the following section.

## 4 NUMERICAL EXAMPLES

In this section, we will first describe the data set. Next, using the training data, we will develop the prediction model, evaluate the gender bias introduced by the prediction model, and illustrate how causal modelling can be used to mitigate the gender bias. Finally, we will detect and mitigate the gender bias using the test data.

### 4.1 Data Set

The data set we use in this study is the Adult data set downloadable from the UC Irvine Machine Learning Repository [14]. The training and test data, respectively, consist of 32,561 and 16,281 records of individuals' age, work class, education, marital status, occupation, race, sex, capital gain, capital loss, hours of work per week, native country and income. Income is the target variable and is binary with two possible values: "<=50K" or ">50K". We will first consider the protected attribute sex.

We will use R to create the prediction model, create the causal model, and evaluate the bias mitigation based on the causal model. The R script we use for this study is presented in the Appendix.

**4.2 Training Data**

In this subsection, we will use the training data to develop and evaluate (1) the prediction model and (2) the bias mitigation model based on causal modeling.

*4.2.1 The Prediction Model*

For simplicity, we use logistic regression as the prediction model and implement it using the *glm* function in R. In practice, other techniques, such as neural networks, can be used to develop the prediction model. Since the prediction model is treated as a black box, we will not present the details of the logistic regression model here.

To generate $\hat{y}$, we use the *predict* function, which yields a probability estimate of an individual's income greater than 50K. Since 76% of the individuals belong to class "<=50K", we use the 76[th] percentile of the prediction $\hat{y}$, which equals to 0.5560, as the threshold for classification. Table 1 presents the results of the classification based on the training data.

Table 1: Classifications Based on $\hat{y}$ (Training)

| | | $\hat{y}$ | |
|---|---|---|---|
| $a$ | $y$ | <=50K (0) | >50K (1) |
| Female (0) | <=50K (0) | **8,571** | 99 |
| Female (0) | >50K (1) | 672 | **440** |
| Male (1) | <=50K (0) | **12,803** | 1,181 |
| Male (1) | >50K (1) | 2,674 | **3,722** |

The numbers in boldface represent correct classifications. The classification accuracy is

$$\frac{8,571+440+12,803+3,722}{8,571+99+672+440+12,803+1,181+2,674+3,722} = 0.8466.$$

The equal opportunity requirement in [9] requires that

$$\Pr\{\hat{Y} = 1 | A = 0, Y = 1\} = \Pr\{\hat{Y} = 1 | A = 1, Y = 1\}.$$

However,

$$\Pr\{\hat{Y} = 1 | A = 0, Y = 1\} = \frac{440}{672+440} = 0.3957,$$

while

$$\Pr\{\hat{Y} = 1 | A = 1, Y = 1\} = \frac{3,722}{2,674+3,722} = 0.5819.$$

The classification accuracy is much lower for female, high-income individuals than for male, high-income individuals, suggesting a lack of equal opportunity.

*4.2.2 The Bias Mitigation Model Based on Causal Modeling*

With sex ($a$), binary income ($y$), and the probability estimate ($\hat{y}$), we solve for the causal model presented in Figure 2 using the *lavaan* package in R. Table 2 presents the path analysis results (for the endogenous variable $\hat{y}$ only, because the results for $y$ are not relevant to bias mitigation).

Table 2: Path Analysis Results

| Coefficient | | Est. | z-value | p |
|---|---|---|---|---|
| Intercept | $\beta_0$ | 0.067 | 31.092 | < 0.001 |
| Sex (a) | $\beta_{a,\hat{y}}$ | 0.117 | 44.446 | < 0.001 |
| Income (y) | $\beta_{y,\hat{y}}$ | 0.415 | 145.250 | < 0.001 |

Since $\beta_{a,\hat{y}}$ is estimated to be significantly different from zero ($z = 44.446$, $p < 0.001$), we can conclude that there is significant gender bias introduced by the prediction model, consistent with the result from the equal opportunity analysis in the previous subsection.

The final causal model is given by

$$\hat{y} = 0.067 + 0.117a + 0.415y.$$

By setting the coefficient of $a$ to zero, we have the corrected prediction $\tilde{y}$ such that

$$\tilde{y} = 0.067 + 0.415y.$$

In other words,

$$\tilde{y} = \hat{y} - 0.117a.$$

We apply the 76th percentile of $\tilde{y}$, which is equal to 0.4537, as the threshold for classification based on $\tilde{y}$. Table 3 presents the results of the classification based on the training data.

Table 3: Classification Based on $\tilde{y}$ (Training)

| | | $\tilde{y}$ | |
|---|---|---|---|
| a | y | <=50K (0) | >50K (1) |
| Female (0) | <=50K (0) | **8,506** | 164 |
| Female (0) | >50K (1) | 571 | **541** |
| Male (1) | <=50K (0) | **12,877** | 1,107 |
| Male (1) | >50K (1) | 2,766 | **3,630** |

Since

$$\Pr\{\tilde{Y} = 1 | A = 0, Y = 1\} = \frac{541}{571+541} = 0.4865,$$

while

$$\Pr\{\tilde{Y}=1|A=1,Y=1\} = \frac{3{,}630}{2{,}766+3{,}630} = 0.5675,$$

we can see that the difference in classification accuracy between female, high-income individuals and male, high-income individuals is now narrowed, suggesting improvement in terms of equal opportunity. Furthermore, the overall classification accuracy is slightly improved to 0.8472.

**4.3 Test Data**

In this subsection, we will use the test data to evaluate the bias mitigation model.

*4.3.1 Applying the Prediction Model.*

We apply the prediction model on the test data to compute $\hat{y}$. After that, we apply the same threshold identified using the training data for $\hat{y}$ (i.e., 0.5560) for classification. Table 4 presents the results of the classification.

Table 4: Classification Based on $\hat{y}$ (Test)

|  |  | $\hat{y}$ | |
| --- | --- | --- | --- |
| a | y | <=50K (0) | >50K (1) |
| Female (0) | <=50K (0) | **4,319** | 37 |
| Female (0) | >50K (1) | 347 | **210** |
| Male (1) | <=50K (0) | **6,398** | 606 |
| Male (1) | >50K (1) | 1,337 | **1,806** |

We again use the equal opportunity criterion to evaluate fairness in the classification. Since

$$\Pr\{\hat{Y}=1|A=0,Y=1\} = \frac{210}{347+210} = 0.3770,$$

while

$$\Pr\{\hat{Y}=1|A=1,Y=1\} = \frac{1{,}806}{1{,}337+1{,}806} = 0.5746,$$

the classification accuracy seems to be much lower for female, high-income individuals than male, high income individuals, suggesting a lack of equal opportunity. The overall classification accuracy is 0.8455.

*4.3.2 Applying the Bias Mitigation Model.*

We apply the same bias mitigation model on the test data to compute $\tilde{y}$. After that, we apply the same threshold identified using the training data for $\tilde{y}$ (i.e., 0.4537) for classification. Table 5 presents the results of the classification.

Table 5: Classification Based on $\tilde{y}$ (Test)

| $a$ | $y$ | $\tilde{y}$ <=50K (0) | $\tilde{y}$ >50K (1) |
|---|---|---|---|
| Female (0) | <=50K (0) | **4,275** | 81 |
| Female (0) | >50K (1) | 288 | **269** |
| Male (1) | <=50K (0) | **6,433** | 571 |
| Male (1) | >50K (1) | 1,385 | **1,758** |

Since

$$\Pr\{\tilde{Y} = 1 | A = 0, Y = 1\} = \frac{269}{288+269} = 0.4829,$$

while

$$\Pr\{\tilde{Y} = 1 | A = 1, Y = 1\} = \frac{1{,}758}{1{,}385+1{,}758} = 0.5593,$$

we can see that the difference in classification accuracy between female, high-income individuals and male, high-income individuals is much narrowed, suggesting improvement in terms of equal opportunity. Furthermore, the overall classification accuracy is slightly improved to 0.8456.

Table 6 summarizes the accuracy performance of the classifications based on the biased and de-biased probability estimates.

Table 6: Comparison of Accuracy Performance

| Data | Estimated Prob. | Classification Threshold | Accuracy Female, High Income | Accuracy Male, High Income | Accuracy Overall |
|---|---|---|---|---|---|
| Train | Biased $\hat{y}$ | 0.5560 | 0.3957 | 0.5819 | 0.8466 |
| Train | De-biased $\tilde{y}$ | 0.4537 | 0.4865 | 0.5675 | 0.8472 |
| Test | Biased $\hat{y}$ | 0.5560 | 0.3770 | 0.5746 | 0.8455 |
| Test | De-biased $\tilde{y}$ | 0.4537 | 0.4829 | 0.5593 | 0.8456 |

## 5 DISCUSSION

To summarize, this paper proposes the use of causal modeling to detect and mitigate bias in the post-processing stage. The proposed method is intuitive and can easily be implemented using the *laavan* package in R. Our results show that when bias exists in the prediction model, causal modeling can be used to detect the bias. In addition, a bias mitigation model can be developed to correct for the bias. Our results show that bias-mitigated classifications show smaller disparity in terms of equal opportunity. Furthermore, the overall classification accuracy may also increase slightly.

An advantage of our statistical approach to bias mitigation is interpretability. Even though an AI model may be a black box, the post-hoc analysis provides a way to describe the nature of bias and the remedies taken to address the bias. This type of explanation can engender trust among different stakeholders in AI. Our approach takes the correlations between protected attributes and the outcome into consideration and provides a justification for deviation from statistical parity.

There is much room for future research on this novel approach to bias mitigation. Firstly, we will apply the same method on problems where $a$ and $y$ are real-valued. Secondly, although we have only addressed bias introduced by the prediction model in this study, we will apply causal modeling to address biases that exist within the data. For example, Madras et al. [11] discussed biases in medical data that affect both the treatment and the outcome. We will extend Figure 2 to incorporate treatment variables in the latent variable models. We will also consider other ways biases could be introduced into AI models, including selection bias discussed in [15].

## APPENDIX: R SCRIPT

```r
1.  library("lavaan")

2.  #read data
3.  train = read.csv("C:\\adult.data.csv")
4.  test = read.csv("C:\\adult.test.csv")

5.  ## 1. Pre-processing
6.  #replace missing values "?" by NA
7.  train[train=="?"] = NA
8.  test[test=="?"] = NA

9.  #following the following example, remove "education", "fnlwgt" and "relationship" from data
10. #https://rstudio-pubs-static.s3.amazonaws.com/235617_51e06fa6c43b47d1b6daca2523b2f9e4.html
11. train$education = NULL
12. train$fnlwgt = NULL
13. train$relationship = NULL
14. test$education = NULL
15. test$fnlwgt = NULL
16. test$relationship = NULL

17. #transform income and sex to numeric 0 or 1
18. train$income[train$income=="<=50K"] = 0
19. train$income[train$income==">50K"] = 1
20. test$income[test$income=="<=50K"] = 0
21. test$income[test$income==">50K"] = 1
22. train$income = as.numeric(train$income)
23. test$income = as.numeric(test$income)
24. train$sex[train$sex=="Female"] = 0
25. train$sex[train$sex=="Male"] = 1
26. test$sex[test$sex=="Female"] = 0
27. test$sex[test$sex=="Male"] = 1
28. train$sex = as.numeric(train$sex)
29. test$sex = as.numeric(test$sex)

30. ## 2. Model Development and Evaluation Using Training Data
```

```
31. #build logistic regression (i.e., the "prediction model")
32. logistic_reg_model = glm(income ~ ., data = train, family = "binomial")

33. #determine threshold rank
34. threshold_rank = sum(train$income==0)

35. #remove "new class" "Never-worked"
36. train$workclass[train$workclass=="Never-worked"] = NA

37. #predict high income propability and perform classification
38. train$predictions = predict(logistic_reg_model, newdata = train, type = "response")
39. threshold = train$predictions[rank(train$predictions) == threshold_rank]
40. train$predictions_class = ifelse(train$predictions > threshold, 1,
ifelse(train$predictions <= threshold, 0, NA))

41. #classification results based on biased predictions (Table 1)
42. table(train$sex, train$income, train$predictions_class)

43. #bias Migation Model Based on Causal Modeling (Table 2)
44. path_model <- '
45. predictions ~ 1 + sex + income
46. income ~ 1 + sex
47. '
48. fit <- sem(path_model, data = train)
49. summary(fit, standardize = FALSE) # unstandardized coefficients are more helpful in
prediction

50. #de-biasing
51. train$debiased_predictions = train$predictions - 0.117*train$sex
52. new_threshold = train$debiased_predictions[rank(train$debiased_predictions) ==
threshold_rank]
53. train$debiased_predictions_class = ifelse(train$debiased_predictions > new_threshold,
1, ifelse(train$debiased_predictions <= new_threshold, 0, NA))

54. #classification results based on de-biased predictions (Table 3)
55. table(train$sex, train$income, train$debiased_predictions_class)

56. ## 3. Evaluation Using Test Data

57. # remove "new class" "Never-worked"
58. test$workclass[test$workclass=="Never-worked"] = NA

59. #predict high income propability and perform classification
60. test$predictions = predict(logistic_reg_model, newdata = test, type = "response")
61. test$predictions_class = ifelse(test$predictions > threshold, 1,
ifelse(test$predictions <= threshold, 0, NA))

62. #classification results based on biased predictions (Table 4)
63. table(test$sex, test$income, test$predictions_class)

64. #de-biasing
65. test$debiased_predictions = test$predictions - 0.117*test$sex
66. test$debiased_predictions_class = ifelse(test$debiased_predictions > new_threshold, 1,
ifelse(test$debiased_predictions <= new_threshold, 0, NA))

67. #classification results based on de-biased predictions (Table 5)
68. table(test$sex, test$income, test$debiased_predictions_class)
```